# The Advantage of Doubling: A Deep Reinforcement Learning Approach to Studying the Double Team in the NBA

Jiaxuan Wang*, Ian Fox*, Jonathan Skaza, Nick Linck, Satinder Singh, Jenna Wiens
University of Michigan, Ann Arbor, MI
*These authors contributed equally to this work

## Abstract

During the 2017 NBA playoffs, Celtics coach Brad Stevens was faced with a difficult decision when defending against the Cavaliers: ``Do you double and risk giving up easy shots, or stay at home and do the best you can?"[1] It's a tough call, but finding a good defensive strategy that effectively incorporates doubling can make all the difference in the NBA. In this paper, we analyze double teaming in the NBA, quantifying the trade-off between risk and reward. Using player trajectory data pertaining to over 643,000 possessions, we identified when the ball handler was double teamed. Given these data and the corresponding outcome (i.e., was the defense successful), we used deep reinforcement learning to estimate the quality of the defensive actions. We present qualitative and quantitative results summarizing our learned defensive strategy for defending. We show that our policy value estimates are predictive of points per possession and win percentage. Overall, the proposed framework represents a step toward a more comprehensive understanding of defensive strategies in the NBA.

## 1  Introduction

In basketball, most defensive metrics focus on discrete and sporadic events, *e.g*., blocked shots, steals, and deflected passes. However, a good defensive play embodies much more than such snapshots capture [2, 7]. For example, throughout a possession, a good defense might force a player into a poor shot location, or force the ball handler to take a series of sub-optimal actions. In this work, we quantitatively measure defensive impact by studying the use and effectiveness of double teaming in the NBA.

When used judiciously, the double team can slow strong offensive players. For example, in Game 4 of the 2015 NBA finals, the Warriors held LeBron to only 20 points and 8 assists by effectively double teaming him. At the same time, doubling against such players can be risky. Doubling one player leaves another player open. How to balance this trade-off is an important question in the NBA and one that can depend on many different factors. According to Andre Iguodala, LeBron's primary defender, "A guy like LeBron who can pass the ball the way he can, you've got to see where his eyes are. If he can see the whole floor, it's tough to double a guy like that."[2] Effective double teaming involves reasoning over the entire court configuration (*e.g.*, where and who the players are) and anticipating the ball handler's next move. One's double teaming strategy must account for the offensive strategy of the opponent.

To characterize the effect of doubling in the NBA, we studied player tracking data from 643,147 possessions. Using a rule-based action detector, we assess how the double team is used by NBA defenses and identify situations when it is most effective. Building on this initial analysis, we propose an approach for learning a defensive strategy for when and where to double team. More specifically, we consider a reinforcement learning (RL) framework, in which we quantify the relationship between the court configuration (*i.e.,* state), the decision to double team and whom to leave open (*i.e.,* action), and the outcome in terms of points per possession (*i.e.,* reward). To deal with the infinite number of court configurations, we use deep RL with a convolutional neural network architecture, we call NothingButNet (NBNet), designed specifically for the task. We train our network to learn a mapping from state and actions to expected cumulative reward. From this mapping, one can construct a policy by selecting the action that maximizes expected reward.

---

[1] cited from http://www.espn.com/nba/story/_/id/19407546/boston-celtics-coach-brad-stevens-says-team-risk-double-teaming-cleveland-cavaliers-lebron-james

[2] cited from http://nba.nbcsports.com/2015/06/13/golden-state-and-the-art-of-double-teaming-lebron-james/

We evaluate the proposed approach on data from the three most recent seasons (including the playoffs). We compare the learned strategy against the actual strategies used. We found that our network's estimate of state-action values had a significant correlation with possession outcomes on a held-out test set ($p<0.001$) and appears to correlate with overall win percentage.

Others have used reinforcement learning to study the effect that field goal attempts have on subsequent 3-pt shot attempts [11], and have used neural networks for play classification [15]. In soccer, recent work has looked at leveraging imitation learning to simulate the outcome of different defensive positions [7]. However, we are the first to use reinforcement learning to study defensive strategies in the NBA. Our use of reinforcement learning (as opposed to supervised or imitation learning) allows us to learn new strategies for double teaming, as opposed to predicting expected outcomes under current play. Our work represents a step toward a more comprehensive understanding and evaluation of defensive play.

The remainder of the paper is organized as follows. Section 2 details the proposed methods including how we define states, actions, and rewards. Section 3 explains our evaluation scheme, including how we use the learned networks to quantify the advantage of double teaming. In addition, Section 3 presents results from the application of the proposed method to three seasons worth of data, summarizing trends across the league, teams and players.

## 2 Methods

We begin with a brief overview of reinforcement learning, introducing definitions and notations used throughout the paper. We then present our proposed method for learning how to effectively double team.

### 2.1 Background and Notation

Here, we briefly review the reinforcement learning (RL) framework; for an in-depth review of RL we refer the reader to [12]. In an RL setting, an agent interacts sequentially with an environment, soliciting a reward. This is commonly modeled using a Markov Decision Process (MDP) $M = (\mathcal{S}, \mathcal{A}, \mathcal{P}, \mathcal{R}, \gamma)$ where $\mathcal{S}$ is the state space, $\mathcal{A}$ is the action space, $\mathcal{P}$ is a transition probability function from state and action to next state, $\mathcal{R}$ is a stochastic function from $\mathcal{S} \times \mathcal{A} \to \mathbb{R}$ and $\gamma \in [0, 1]$ is the discount factor for the reward. In an episodic setting, an agent observes the current state $s_t \in \mathcal{S}$, chooses an action $a_t \in \mathcal{A}$, and then transitions to $s_{t+1}$, according to some probability distribution $\mathcal{P}_{s_t}^a$. In addition, the agent receives an instantaneous reward, $r_t \coloneqq \mathcal{R}(s_t, a_t)$. This process continues until reaching a terminal state at time step $T$ (*i.e.,* the end of the episode).

In this setting, the agent aims to maximize the expected value of cumulative discounted reward $G_t \coloneqq \sum_{t=t_0}^{T} \gamma^{t-t_0} r_t$. An agent behaves according to some policy $\pi$ where $\pi(a|s) \coloneqq \mathcal{P}_{s_t}^a$ The optimal state-action value function $Q^{\pi^*}(s,a)$ is the expected cumulative reward of starting in state $s$, executing action $a$, and then following the optimal policy $\pi^*$. Formally, $Q^{\pi^*}(s,a) = \max_\pi \mathbb{E}_{a_{t+1:T-1} \sim \pi} [G_t | s_t = s, a_t = a]$. In addition to state-action value, we define optimal state value $V^{\pi^*}(s) = max_a Q^{\pi^*}(s,a)$. Finally, we define the state-action advantage as $A^{\pi^*}(s,a) = Q^{\pi^*}(s,a) - V^{\pi^*}(s)$, or the expected difference in cumulative reward had $a$ been selected.

In the next section, we explain how we use this framework to find the optimal policy for double teaming. We begin by defining the set of actions $\mathcal{A}$, our state representation $\mathcal{S}$, and the reward $\mathcal{R}$.

## 2.2 Double Teaming with an RL Framework

Applied to player trajectory data, the RL framework considers each possession as an episode. The episode begins once all players have crossed half court and concludes when the shot clock resets. We discretize these episodes into 1 second windows. From the defense's perspective, at each second, the team (*i.e.*, agent) must make a decision (*e.g.*, to double team the ball handler or not). This decision can depend on many factors including the status of the game, which players are on the court, and player locations. Ultimately, this sequence of decisions results in an outcome determining the total cumulative reward (*e.g.*, a blocked shot). Intuitively, if the defense makes a series of poor or sub-optimal decisions, the offense will score, and the resulting reward will be lower than if the defense had made better decisions. Below, we provide additional details regarding how we i) define and detect actions, ii) handle the continuous state space and iii) measure reward.

### 2.2.1 Action Space and Action Detector

While the defense is faced with myriad decisions during play, we focus on the decision of whether or not to double team the ball handler. We define a discrete action space based on the location of the open player. We discretize the court into 19 different regions, in keeping with [1] (**Figure 1**). Given this court discretization and the location of players on the court, we label each 1-second window with one of 20 possible actions. The defense can either decide to stay, or double team the ball handler. Since the player the defense chooses to leave open can be in any one of the 19 regions, this presents 20 possible actions in total. However, for any given state, only a subset of actions are valid, since it is infeasible to leave an open man in a region not occupied by an offensive player. Thus, the set of feasible actions at each time $\mathcal{A}_t \subset \mathcal{A}$ depends upon $s_t$, where $a \in \mathcal{A}_t$ if and only if $a$ is the null action (no double teaming) or $a$ corresponds to leaving an open man in a region occupied by an offensive player (excluding the ball handler) in $s_t$.

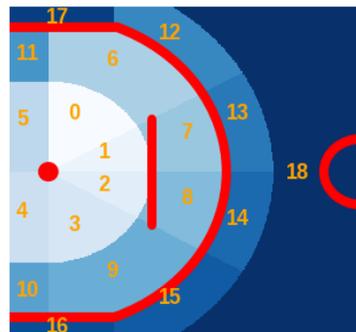

**Figure 1**. We discretize the half court into 19 distinct regions.

To detect the action of double teaming in the data, we developed the simple rule-based classifier described below. We tried using the `Who's Guarding Whom' system to detect double teaming [2], but ultimately found that an action detector tailored to the specific problem of double teaming was more accurate. The classifier looks for the presence of at least two defensive players within a radius of the ball handler, while accounting for the possibility that two offensive players are close to one another, bringing the defense close (but not double teaming). We summarize this rule in **Figure 2**. To ensure that this simple rule could accurately capture double teaming, we compared its annotations with those given by two humans on a random subset of 100 possessions. We found that the action detector performed within the level of inter-rater agreement, see **Figure 2**(c), demonstrating that it is a reasonable approach to detecting double teams.

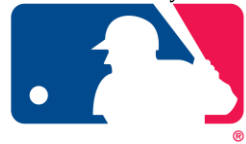

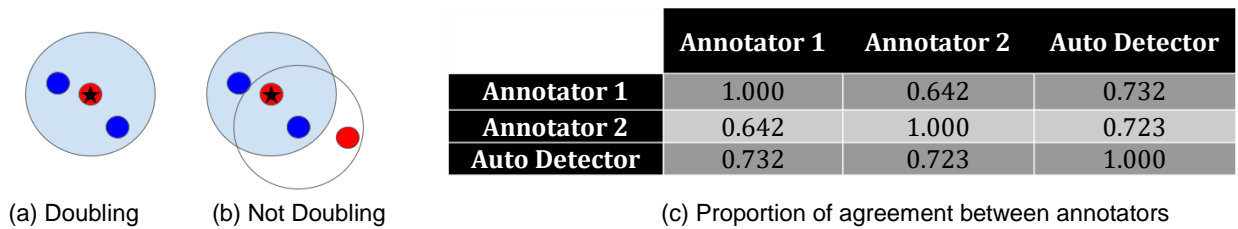

(a) Doubling  (b) Not Doubling

|  | Annotator 1 | Annotator 2 | Auto Detector |
|---|---|---|---|
| **Annotator 1** | 1.000 | 0.642 | 0.732 |
| **Annotator 2** | 0.642 | 1.000 | 0.723 |
| **Auto Detector** | 0.732 | 0.723 | 1.000 |

(c) Proportion of agreement between annotators

**Figure 2.** Identifying double teams. In (a), we find two defensive players within the defensive radius of the starred player, removed from the other offensive players → doubling. In (b) one of the defensive players is within the defensive radius of another offensive player → not doubling. In (c) we see that the model agrees with the human annotators more often (72.3% and 73.2%) than the humans agree with each other (64.2%).

When labeling possessions based on the presence of double teaming, we consider those possessions in which the defense selects a ``double team'' action for at least two consecutive 1-second windows as double teaming. Note that this does not necessarily mean the defense was doubling for two seconds, just that the action occurred over the course of two windows. This eliminates short-lived ``double teams'' that occur during screens or during drives to the basket.

### 2.2.2   State Representation

The decision to double team or not can depend on many factors. We try to account for as many as possible by considering a continuous state representation that encapsulates player trajectories, player heights, weights, shooting abilities, current state of the game, shot clock, and game clock. As input to our network, we use both images, as in [3], and flat features.

Our image representation includes three types of channels: i) one court channel encoding the region number, as defined in **Figure 1**, of each pixel, ii) 11 trajectory channels (for the 10 players and ball), and iii) five offensive player shooting percentage channels, each of size 47×50 in the pixel space (*i.e.*, the half court discretized by square feet). Building upon work by Harmon *et al.*, we convert the player trajectories, from their original $(x, y)$-coordinate format to an image representation. For each player on the court, we build an image of his trajectory over a 1-second window. The pixel value for a player location exponentially decreases going back in time, directly encoding temporal information into the image. Finally, we include an additional image channel for each offensive player on the court, capturing their shooting percentages across different regions. When estimating shooting percentages, we use data up to but not including the current game so as to respect the causal ordering of events. This results in a sparse 17 channel image (as opposed to a standard 3 channel RGB image). The channels are sorted across images by team and position within a team to preserve image semantics across examples (*i.e.*, the 1st channel contains the trajectory of the offensive center if one is fielded). A sample trajectory is given in **Figure 3**.

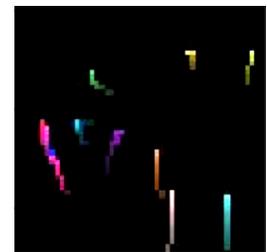

**Figure 3.** Trajectory over 1 second of play. Each color represents a unique channel.

In addition to these image inputs, we include flat features that do not have a spatial component. These flat features pertain to the shot clock, game clock, quarter, and weight and height of each player on the court. Again, we order the player features according to player team and position. After binning continuous values based on quintiles, the resulting feature vector consisted of 93 flat features. These were fed into the model along with the image channel input (**Figure 4**).

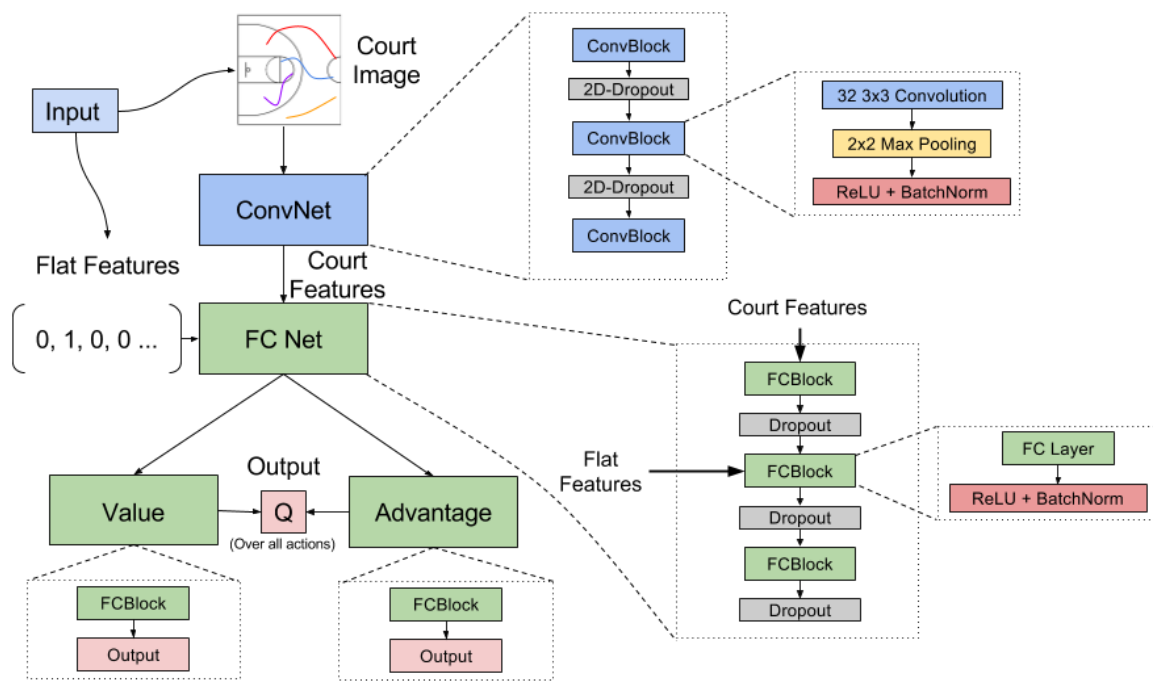

**Figure 4.** NBNet architecture. We use a convolution neural network (ConvNet) to extract visual features from the court together with a fully connected network to incorporate flat features. The final hidden layers are split into value and advantage streams that output $V(s)$ and $A(s,a)$ respectively, with output $Q(s,a) = V(s) + A(s,a)$. FC: fully connected

### 2.2.3 Reward

We consider a reward signal based on the points obtained over a possession. In our setting, only the final second of a possession is associated with non-zero reward. By removing the discount factor, the total cumulative reward at any point during the possession is equal to the terminal reward. If a foul occurs, we include the result of the free throws in our reward. Because we consider each possession from the perspective of the defense, the reward equals the negative points scored and takes on values in the set $\{0, -1, \ldots, -5\}$, the closer to 0, the better.

## 2.3 Model Architecture and Training

Given the RL framework discussed above, we train a state-action value estimator to learn a mapping $s \to Q^\pi(s,a)$ for all actions $a \in \mathcal{A}$. We use a dueling convolutional neural net as our estimator [16]. Such architectures natively separate state-value and action advantage. They have been demonstrated to be a competitive architecture for vision-based control [5].

State-action networks learn the Q-values for each action $a_t$ given a state $s_t$, at some particular time step $t$ by minimizing the temporal difference error across the $N$ episodes in the training set $\{s_1^i, a_1^i, r_1^i, \ldots, s_{T_i-1}^i, a_{T_i-1}^i, r_{T_i-1}^i, s_{T_i}^i\}_{i=1}^N$:

$$\mathcal{L} = \sum_{i=1}^{N} \sum_{t=1}^{T_i-1} [Q^{\pi^*}(s_t^i, a_t^i) - (r_t^i + V^{\pi^*}(s_{t+1}^i))]^2$$

Where $\pi^*$ represents the policy, and both $Q^{\pi^*}$ and $V^{\pi^*}$ are estimated using the network (though to improve stability we estimate $V^{\pi^*}$ using a periodically cached version of our network as in [10]). Given the training data, we learn a state-action network using a policy that is greedy with respect to the value estimates:

$$\pi^*(a|s_t) = \begin{cases} 1, & if\ a = \underset{a \in \mathcal{A}_t}{\mathrm{argmax}}\ Q^{\pi^*}(s_t, a) \\ 0, & otherwise \end{cases}$$

$$V^{\pi^*}(s_{t+1}) = \max_{a\ \in \mathcal{A}_{t+1}} Q^{\pi^*}(s_{t+1}, a)$$

The feasible action set $\mathcal{A}_t$ is used both for policy predictions and value estimation. This approach of policy learning via value estimation is called Q-learning. Given unlimited exploratory training data, such a policy is provably optimal for any finite size MDP [17]. However, our use of historical data means we cannot sufficiently explore the state-action space to make such guarantees. Training deep neural networks with Q-learning was first shown to be viable for vision-based control in [10]. To improve the Q-value estimates, we use double Q-learning as described in [14]. An overview of our full NBNet architecture is given in **Figure 4**. Our model implementation and training code is available online[3].

## 3   Evaluation and Results

We trained and evaluated on data collected from the SportVU optical tracking dataset augmented with play-by-play data. These data have been previously described by [8,9,18]. We considered data from the three most recent seasons, totaling 875,412 possessions. For our analysis, we excluded possessions in which not all of the players cross half-court, since these represent transition plays in which the use of double teaming may differ. This results in a final set of 643,147 possessions. In our first set of results, we summarize the application of our action detector to the data. We identify how often teams use the double team, and when and where it appears most effective. We then move on to present an in-depth analysis of the learned value function, demonstrating the promise of an observational RL framework for value estimation.

### 3.1   Observational Analysis

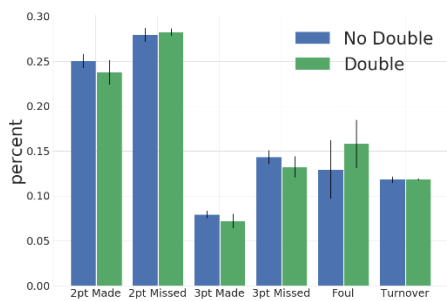

**Figure 5.** Relative frequencies of outcomes for possessions in which the ball handler was and was not double teamed. Error bars correspond to empirical 95% confidence intervals.

We labeled each of the 643,147 possessions as *Double* if we detected that the defense double teamed the ball handler (based on our action detector, Section 2.2.1) for at least 2 consecutive seconds, and *No Double* otherwise. In total, 4.8% were labeled as *Double*. This fraction varied across teams from 3.8% (Portland) to 6.7% (Milwaukee). The majority of teams tend to double team the ball handler between 4-5% of possessions.

After labeling each possession, we grouped possessions by outcome. Each possession fell into one of six possible categories: 2pt made/missed, 3pt made/missed, foul, or turnover. Double teaming is inherently a high-risk, high-reward decision. The distribution over outcomes associated with double teaming reflects this trade off (**Figure 5**). Overall, double teaming results in a significantly lower field goal percentage for the offense. However, this comes at the expense of a significantly greater likelihood of the possession ending in a foul.

---

[3] https://github.com/igfox/AdvantageOfDoubling

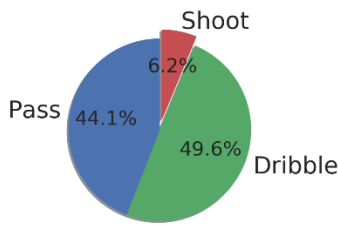

Next, we looked at the data from the perspective of the ball handler, investigating what actions were taken immediately following a double team and the outcome. Across the NBA, when double teamed ball handlers predominantly tend to pass or dribble the ball, shooting only 6.2% of the time when double teamed (**Figure 6**).

**Figure 6.** Once double teamed, a player can choose to either pass, shoot, or dribble the ball, but most choose to pass

To assess which offensive players perform best when double teamed, we measured a team's overall points per possession when a given player was on the court and compared this to the team's points per possession (ppp) when that given player was double teamed. Here, we restricted our analysis to players who faced at least 150 double teams. Among guards, John Wall emerged as the most effective player against the double team. On average, Wall's team scored 1.07ppp when he was double teamed versus 0.89ppp when he was not. This increase may reflect Wall's ability to effectively utilize the open man. Lou Williams is the guard most negatively affected by the double team (0.68ppp *vs*. 0.95ppp). Among forwards, Rudy Gay comes up as most effective (0.86 ppp *vs*. 0.98ppp). On the other end of the spectrum, Kevin Durant is most negatively affected (0.99 ppp *vs*. 0.90 ppp).

In **Figure 7,** we plot the average ppp for each of the above players, categorized by the decision the offensive player made once they were double teamed. It appears more beneficial for the player to pass the ball rather than keep the ball themselves. Despite this fact, players do not always make this decision as seen in **Figure 6**.

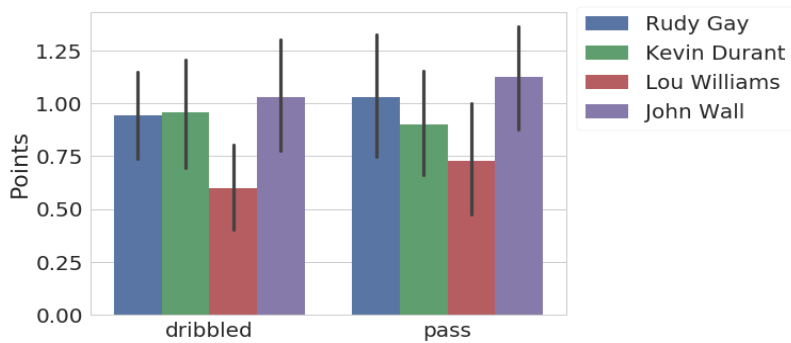

**Figure 7.** Expected points per possession when players decide to dribble or pass when double teamed.

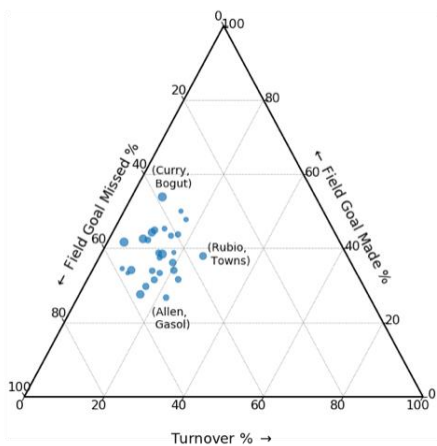

**Figure 8.** Relationship between Field Goals, Turnovers, and Fouls (indicated by dot size) for tandems with at least 50 double teams.

In addition to considering what the offense does when faced with a double team, we also looked at how the defense fares. We examined all tandems with at least 50 double teams over the last three seasons. **Figure 8** illustrates the relationship among field goals, fouls committed, and turnovers forced for each of these pairs. The pairing of Kyle Lowry and Jonas Valanciunas led all pairings in terms of fewest points allowed per possession (0.64), followed by Chris Paul and DeAndre Jordan (0.70) and Klay Thompson and Draymond Green (0.74). The tandem of Ricky Rubio and Karl-Anthony Towns was most proficient at forcing turnovers out of the double team--21.4% of their double teams resulted in a turnover.

The results from this empirical analysis begin to shed light on the tough decisions teams must make with respect to when and whom to double team. In the next section, we take a closer look at the value behind the decisions teams are making, while controlling for additional factors through an RL approach.

## 3.2 Reinforcement Learning Analysis

In this section, to control for the effect of the offensive team, we consider how teams defend against a single team. We focus on the Cleveland Cavaliers, since the Cavs i) had a lot of data (so that we could learn a policy), ii) had a relatively stable roster (so that its offensive strategy did not vary too much) and iii) are strong offensively (so that good defense is necessary).

Focusing on just offensive possessions for the Cleveland Cavaliers resulted in 22,695 possessions. After splitting the data randomly on possessions, we trained the network on 70% of the data, validated our results on 10%, and performed our final evaluations on the remaining 20%. Applied to a held-out possession $\tau_i = (s_1^i, a_1^i, r_1^i, s_2^i, a_2^i, r_2^i, \ldots, s_{T_i}^i)$, at each time-step, $t$, the learned network estimates the Q-values, $Q^{\pi^*}(s_t^i, a)$ for all $a \in \mathcal{A}_t$ for each of the feasible actions given the state representation $s_t^i$.

The quantitative evaluation of policies on data collected off-policy is challenging. While several recent advances have been made in methods for off-policy evaluation [6, 11], they require that the behavior policy have a known distribution. This does not hold in our scenario, and assuming a deterministic behavior policy would severely limit our evaluation data (requiring that we evaluate only on trajectories that already follow $\pi^*$). Since we do not correct for changes to the data distribution generated under the learned policy, the Q-values we report when following $\pi^*$ are biased [6, 13]. Despite this fact, we still observe that our learned policy correlates with an increased expected reward. 19 of the 20 actions correspond to double teaming. Because of this, the learned policy suggests double teaming far more often than not (in 90.6% of possessions). To mitigate this effect, we suppressed the selection of a double team action unless its Q-value exceeded that of the man-to-man action by at least 0.2. We selected this threshold using the validation data, verifying that the average Q-values still correlated with observed outcomes. Once we applied this post-processing step, the learned policy suggested double teaming in 29.29% of possessions. This is closer to the observed double teaming ratio of 33.92% when double teaming is defined on a per-second basis.

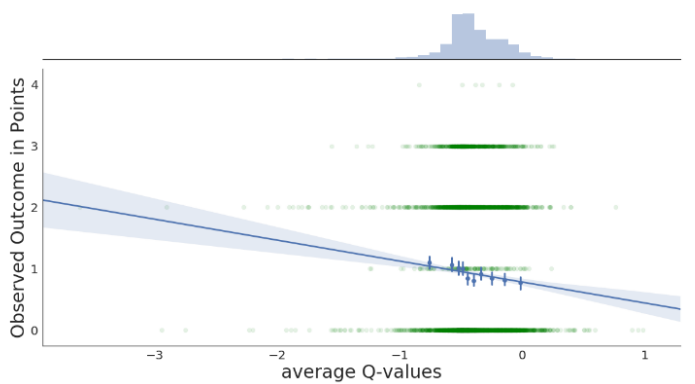

**Figure 9.** We look at the estimated Q-values averaged over the course of a possession compared to the observed number of points scored at the end of the possession. As expected, we see a downward trend where a defensive `stop' corresponds to higher Q-values. This has correlation -0.08 ($p < 0.001$)

**How predictive are the Q-values?** Given a set of 4,482 held-out possessions, we evaluated the accuracy of the learned network by comparing the estimated Q-values associated with the action taken at each time step $t$, $Q^{\pi^*}(s_t, a_t)$, to the observed outcome (*i.e.*, points scored by the offense). That is, we computed $q_{avg}(i) = \frac{1}{T_i}\sum_{t=1}^{T_i} Q^{\pi^*}(s_t^i, a_t^i)$ for each possession $i$. If the network is a good estimator, then a higher $q_{avg}$ should be associated with a lower number of points scored (since this represents a good outcome for the defense). In **Figure 9** we plot $q_{avg}$ against the observed outcome across possessions in the test set. For clarity, we binned the averaged Q-values into deciles and report the average score within each decile. Note that the regression line is fit to the underlying data, not the bin values. We observe the expected downward trend. That is, a higher average Q-value is associated with fewer points scored (*i.e.*, a good defensive outcome). We also observe a tight clustering of mean Q-values within the range -1 to 0. This indicates that our Q-values are positively biased relative to the expected cumulative reward, as the average reward is around -1.4. This upward bias of Q-values is a well-known problem in Q-learning [4]. Despite this bias, the Q-values are predictive of reward, a highly encouraging result. In addition, we observed that the Q-values are associated with higher-level measures of team performance. **Figure 12** shows a positive correlation between team-wise $q_{avg}$ and win percentage against the Cavs.

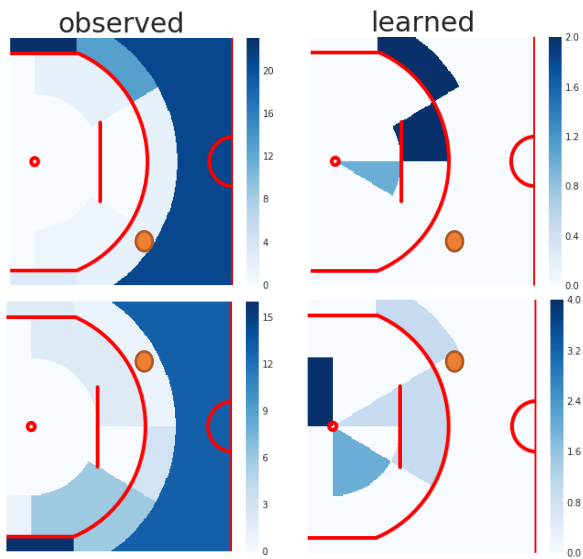

**Figure 10.** Observed (left figures) and learned policies (right figures) for double teaming Kyrie Irving at the 3-point line. The top figures depict possessions when Kyrie is at the left wing of the 3 point line, while the bottom figures depict cases when Kyrie is at the right wing of the 3 point line. The observed and learned policies differ significantly. The observed policy demonstrates many open players are left in the back, while the learned policy suggests the extra defender should come from the inner court. Both the observed and learned policies demonstrate significant asymmetry, showing that it's better to leave an open player away from the ball handler

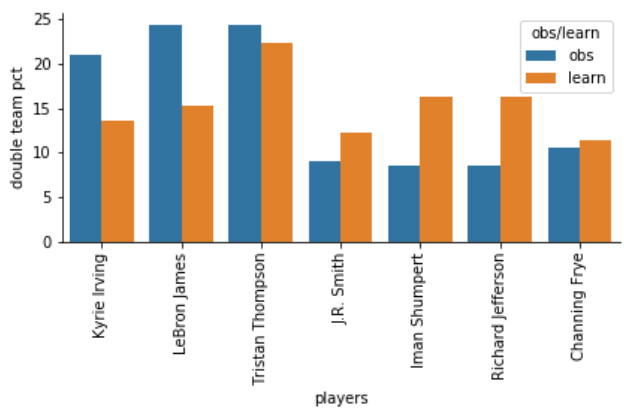

**Figure 11.** Percentage of time we observe teams double teaming individual players (obs) and percentage of time our learned policy suggests double teaming those same players (learn).

**Visualizing the Learned Policy.** Once we established that the estimates for the Q-values were reasonable, we performed both qualitative and quantitative analyses of the learned policy. In particular, again on held-out data, we compared the distribution over actions suggested by the learned policy $\pi^*$ to the empirical distribution of observed actions. That is, for each observed state and action pair, $(s_t^i, a_t^i)$, we compared how $a_t^i$ differs from $\underset{a}{\mathrm{argmax}}\, Q^{\pi^*}(s_t, a)$.

To try to get at the question of how a team might improve their double teaming strategy, we focus on double teaming Kyrie Irving. We looked at all instances in which Kyrie had the ball around the perimeter (at least 15 feet from the basket). For each time step that meets these criteria, we considered the observed action versus the action suggested by $\pi^*$. For those cases in which the defense decided to double team Kyrie, **Figure 10** shows the locations of the open players (left), and where the open player should have been had the teams been acting according to $\pi^*$ (right). We observe the counter-intuitive result that it is better to leave an open man in the paint than in the back. This could suggest that it's important for the double teamer to position himself such that he blocks a potential pass to the open man. The fact that the network learns to capitalize on the asymmetry of the situations, as evidenced by values within the paint, indicates the policy is responding to positional information in $s_t$.

In addition to where to leave the open man, we can look how often we should be double teaming certain players. **Figure 11** shows differences in double teaming trends across players with the observed versus learned policy. Our learned approach is more hesitant to double team star players, and is more likely to double team role players.

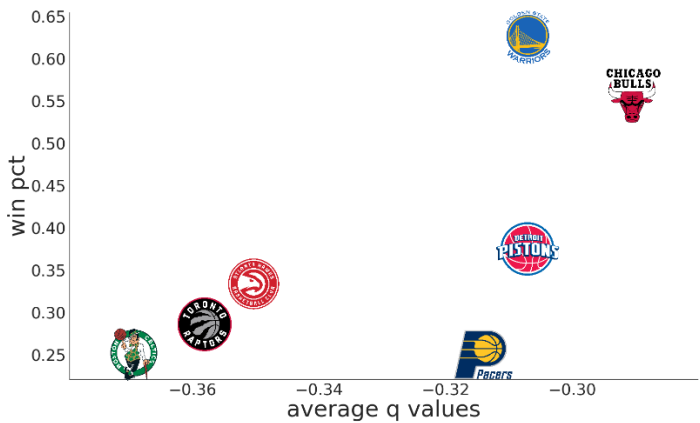

**Figure 12.** An analysis of each team's empirical performance, as measured by win percentage, against their performance as evaluated by $Q^{\pi^*}$. We see a strong relationship between the measures, indicating the reliability of $Q^{\pi^*}$

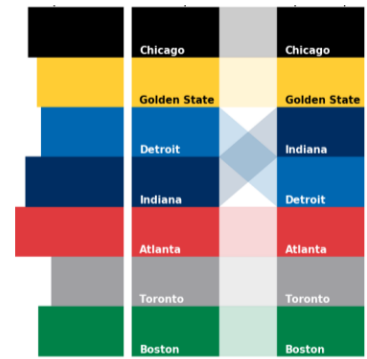

**Figure 13.** The performance and potential of teams as measured by $Q^{\pi^*}(s_t, a_t)$. The leftmost bars indicate advantage, the left ranking is done on Q-values, the right is done on potential under optimal play ($V^{\pi^*}$)

**Who's the best (and who could improve)?** The average Q-value associated with the trajectory of states and actions for a possession may be high due to either the value of the state, or the advantage conferred by the selected actions. For example, if none of the players on the Cavs' starting line-up are on the court, then the Q-value will be high since the team is less likely to score. Thus, we separate out the advantage from the Q-value, shown in **Figure 13.** The ranking in the plot corresponds to the average $q_{avg}$ for each team. We only consider teams who played the Cavs in the playoffs, since this provides more data to train and evaluate on. We sort the graph from the bottom to the top (the higher the better) using $q_{avg}$ over all relevant possessions. The left-most bar corresponds to the magnitude of average advantage for each team (*i.e.*, $A^{\pi^*}(s_t, a_t)$). The right ranking is done using the theoretical optimal performance of the teams ($V^{\pi^*}(s_t)$). From this plot, we observe that Chicago and Golden State are currently best at defending against the Cavs, while Indiana has great potential under "optimal" play. Notably, all teams can improve their defense under the learned policy.

## 4 Conclusion

In this paper, we developed a framework for studying the double team in the NBA. Using a rule-based action detector, we labeled hundreds of thousands of possessions from the last three seasons as either containing a double team or not. We then applied a deep reinforcement learning approach to learn a mapping from court/game configuration to the action that minimizes the number of points scored by the offense.

There are several limitations to our analysis due mostly to the fact that we only have observational data. First, $\pi^*$ learned in this work is optimal only with respect to the state-action pairs explored in the data. Second, while it is typical to evaluate learned policies through application or a simulator, we are limited to evaluating on observed state-action pairs. Third, while we pooled data from across three seasons, there is still a limited number of observations for both training and evaluation purposes. Finally, we rely on a rule-based action detector for identifying double teams and are thus limited by the accuracy of this detector. Future work could improve the action-detector which would in turn impact the RL results.

This work represents the first time that defensive strategy in the NBA has been analyzed using an RL framework. Though preliminary, it demonstrates the potential for algorithmic analyses of the types of problems plaguing coaches. While we chose to focus on doubling, the proposed framework generalizes beyond this specific task. By modifying the defined action space, the same approach could be used to answer other questions about offensive and defensive strategy. Going forward, the proposed approach could be applied to player tracking data across several different settings to study sequential decision making in the NBA.

## References

[1] Chang, Yu-Han, et al. "Quantifying shot quality in the NBA." *Proceedings of the 8th Annual MIT Sloan Sports Analytics Conference. MIT, Boston, MA*. 2014.

[2] Franks, Alexander, et al. "Counterpoints: Advanced defensive metrics for NBA basketball." *Proceedings of the 9th Annual MIT Sloan Sports Analytics Conference, Boston, MA*. 2015.

[3] Harmon, Mark, Patrick Lucey, and Diego Klabjan. "Predicting Shot Making in Basketball using Convolutional Neural Networks Learnt from Adversarial Multiagent Trajectories." *arXiv preprint arXiv:1609.04849* (2016).

[4] Hasselt, Hado V. "Double Q-learning." *Advances in Neural Information Processing Systems*. 2010.

[5] Hessel, Matteo, et al. "Rainbow: Combining Improvements in Deep Reinforcement Learning." *arXiv preprint arXiv:1710.02298* (2017).

[6] Jiang, Nan, and Lihong Li. "Doubly robust off-policy value evaluation for reinforcement learning." *arXiv preprint arXiv:1511.03722* (2015).

[7] Le, Hoang M., et al. "Data-driven ghosting using deep imitation learning." *Proceedings of the 11th Annual MIT Sloan Sports Analytics Conference. MIT, Boston, MA*. 2017.[8] McIntyre, Avery et al.,

"Recognizing and Analyzing Ball Screen Defense in the NBA", *Proceedings of the 10th Annual MIT Sloan Sports Analytics Conference. MIT, Boston, MA*. 2016.

[9] McQueen, Armand et al. "Automatically Recognizing On-Ball Screens." *Proceedings of the 8th Annual MIT Sloan Sports Analytics Conference. MIT, Boston, MA*. 2014.

[10] Mnih, Volodymyr, et al. "Human-level control through deep reinforcement learning." *Nature* 518.7540 (2015): 529-533.

[11] Neiman, Tal, and Yonatan Loewenstein. "Reinforcement learning in professional basketball players." *Nature communications* 2 (2011): 569.

[12] Sutton, Richard S., and Andrew G. Barto. *Reinforcement learning: An introduction*. Vol. 1. No. 1. Cambridge: MIT press, 1998.

[13] Thomas, Philip, and Emma Brunskill. "Data-efficient off-policy policy evaluation for reinforcement learning." *International Conference on Machine Learning*. 2016.

[14] Van Hasselt, Hado, Arthur Guez, and David Silver. "Deep Reinforcement Learning with Double Q-Learning." *AAAI*. 2016.

[15] Wang, Kuan-Chieh, and Richard Zemel. "Classifying NBA offensive plays using neural networks." *Proc. MIT SLOAN Sports Analytics Conf.* 2016.

[16] Wang, Ziyu, et al. "Dueling network architectures for deep reinforcement learning." *arXiv preprint arXiv:1511.06581*(2015).

[17] Watkins, Christopher JCH, and Peter Dayan. "Q-learning." *Machine learning* 8.3-4 (1992): 279-292.

[18] Wiens, Jenna, et al. "To crash or not to crash: A quantitative look at the relationship between offensive rebounding and transition defense in the NBA." *Proceedings of the 6th Annual MIT Sloan Sports Analytics Conference. MIT, Boston, MA*. 2013.
122018 Research Papers Competition
Presented by: